\providecommand{\paperversion}{arxiv}

\newif\ifarxiv
\def\UniDexArxivMode{arxiv}
\def\UniDexIcraMode{icra}
\edef\UniDexSelectedMode{\paperversion}
\ifx\UniDexSelectedMode\UniDexArxivMode
  \arxivtrue
\else
  \ifx\UniDexSelectedMode\UniDexIcraMode
    \arxivfalse
  \else
    \PackageError{UniDex-ViTac}{Invalid paperversion: \paperversion}%
      {Use exactly icra or arxiv.}
  \fi
\fi

\documentclass[letterpaper,10pt,conference]{ieeeconf}
\IEEEoverridecommandlockouts
\usepackage[warnundef]{Styles/jabbrv} 

\usepackage{graphicx}
\usepackage{caption}
\usepackage{subcaption}
\usepackage{amsmath,amssymb}
\usepackage{multirow}
\usepackage{booktabs}
\usepackage{siunitx}
\usepackage{cite}
\usepackage{array}
\usepackage{makecell}
\usepackage{tabularx}
\usepackage{mathtools}
\usepackage{microtype}
\usepackage{url}
\usepackage{xcolor}
\usepackage{etoolbox}
\makeatletter
\@ifpackageloaded{natbib}{}{\let\NAT@parse\undefined}
\makeatother
\usepackage{hyperref}
\definecolor{figgreen}{HTML}{17865E}
\definecolor{figblue}{HTML}{2867D1}
\definecolor{figpink}{HTML}{C12F9B}
\definecolor{linkblue}{RGB}{0,102,204}
\definecolor{referenceblue}{HTML}{0066CC}

\colorlet{citationcolor}{referenceblue}
\colorlet{bibliographynumbercolor}{black}

\makeatletter
\renewcommand{\@biblabel}[1]{\textcolor{bibliographynumbercolor}{[#1]}}
\makeatother

\AtBeginEnvironment{thebibliography}{%
  \color{black}%
  \hypersetup{urlcolor=linkblue,linkcolor=linkblue}%
}
\hypersetup{
  colorlinks=true,urlcolor=linkblue,linkcolor=referenceblue,citecolor=citationcolor,
  pdftitle={UniDex-ViTac: Learning Unified Visuo-Tactile Dexterous Manipulation Policy from Human Video Data},
  pdfsubject={},pdfkeywords={},pdfcreator={},pdfproducer={}
}
\ifarxiv
  \hypersetup{pdfauthor={Hyesung Lee, Si-Hwan Heo, Sungwook Yang}}
\else
  \hypersetup{pdfauthor={}}
\fi

\newcommand{\norm}[1]{\left\lVert #1 \right\rVert}

\title{\LARGE \bf
UniDex-ViTac: Learning Unified Visuo-Tactile Dexterous Manipulation Policy from Human Video Data
}
\ifarxiv
\author{Hyesung Lee, Si-Hwan Heo and Sungwook Yang \\
  \\
\thanks{This work was supported by the National Research Foundation of Korea (NRF) grants funded by the Korean government (MSIT) (No. RS-2024-00464386, No. RS-2025-25396144), and by the Korea Institute of Science and Technology (KIST) Institutional Program.}%
\thanks{H. Lee, S.-H. Heo, and S. Yang are with the Center for Humanoid Research, Korea Institute of Science and Technology, Seoul 02792, Republic of Korea. H. Lee is also with the Kim Jaechul Graduate School of AI, Korea Advanced Institute of Science and Technology, Seoul 02455, Republic of Korea. (\texttt{\{hs981002, hershey, swyang\}@kist.re.kr}).}%
}
\else
\author{Anonymous Authors}
\fi

\begin{document}
\maketitle
\thispagestyle{empty}
\pagestyle{empty}

\begin{abstract}
Human videos provide demonstrations of dexterous manipulation but lack robot-executable actions and tactile measurements. We present \textbf{UniDex-ViTac}, a framework that uses human-video-guided simulation to generate robot demonstrations paired with fingertip contact observations for training a deployable visuo-tactile policy. Object-specific residual reinforcement learning specialists adapt annotated human--object interaction references to a robotic arm--hand system. Their successful rollouts pair final robot action targets with robot-side fingertip contact observations. From 50 human demonstrations across ten objects, we collect 10,000 simulated trajectories to train a single Action Chunking with Transformers (ACT) based generalist. The policy combines point clouds, proprioception, and four binary contact signals encoded through fingertip labels and a separate token, without requiring human references or privileged object identity and pose at deployment. The contact-augmented configuration achieves 68.3\% macro-average success in simulation, compared with 55.5\% for the point-cloud-only baseline. Without real-robot demonstrations or policy fine-tuning, it succeeds in 73/110 physical trials (66.4\%) across six seen and five unseen objects, compared with 60/110 (54.5\%) for the baseline, an increase of 11.8 percentage points. These results support the feasibility of learning a unified visuo-tactile dexterous manipulation policy from video-guided simulated interactions.
Project page: \url{https://unidex-vitac.github.io/}.
\end{abstract}

\section{Introduction}
\label{sec:intro}

Imitation learning has driven significant progress in robotic manipulation using human demonstrations~\cite{chi2023diffusion,zhao2023learning,mandlekar2021matters}. However, scaling this paradigm to multi-fingered dexterous hands remains challenging, as collecting demonstrations requires complex teleoperation interfaces to bridge embodiment gaps~\cite{handa2020dexpilot,qin2023anyteleop}. While human interaction videos offer demonstrations without robot teleoperation, leveraging them for contact-aware robotic policies presents two fundamental challenges: human movements do not directly translate to executable robot kinematics, and passive video recordings lack tactile or contact measurements.

Although geometric human--object references can guide initial motion trajectories~\cite{qin2022dexmv,singh2024hop}, they do not by themselves ensure that a physical robot can grasp and lift objects under its own kinematic and contact dynamics. Vision provides geometric context, whereas tactile observations indicate local physical contact during execution. This raises a key question: \emph{How can human videos without tactile measurements support learning a deployable, contact-conditioned robot policy?}

To bridge this gap, we use simulated robot interactions to generate executable action targets paired with contact observations. Human-derived motion references guide object-specific residual reinforcement learning (RL) specialists in adapting demonstrations to the target robot. Successful rollouts pair the resulting robot motions with contact signals generated by simulated interaction dynamics. These rollouts train a single generalist policy for reference-free execution.

\begin{figure*}[t]
    \centering
    \includegraphics[width=\textwidth]{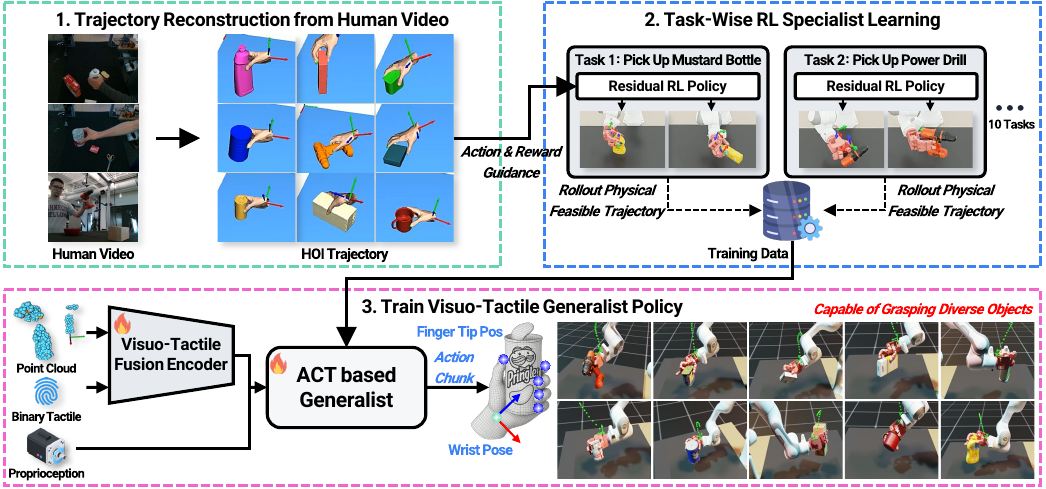}
    \caption{Overview of UniDex-ViTac. (1) Annotated human videos provide human--object interaction (HOI) references (\textcolor{figgreen}{green}). (2) Object-specific residual RL specialists adapt these references in simulation and generate robot action--contact demonstrations (\textcolor{figblue}{blue}). (3) The pooled demonstrations train a single ACT-based~\cite{zhao2023learning} visuo-tactile generalist (\textcolor{figpink}{pink}). At deployment, the generalist uses point clouds, proprioception, and four contact bits, without human references or privileged object identity or pose.}
    \label{fig:figure_1}
\end{figure*}

As illustrated in Fig.~\ref{fig:figure_1}, \textbf{UniDex-ViTac} links this data generation pipeline to an Action Chunking with Transformers (ACT) based architecture~\cite{zhao2023learning}. We incorporate contact signals via a four-bit fingertip interface: each bit labels the corresponding forward-kinematics (FK) fingertip point, while the full four-bit vector is encoded as a separate contact token. These two paths provide spatial anchors and a direct representation of the contact pattern. At test time, the policy operates directly on point clouds, proprioception, and binary contact inputs, without requiring object identities, ground-truth object poses, or reference trajectories.

UniDex-ViTac builds upon video-guided data generation~\cite{chen2024vividex}, specialist-to-generalist distillation~\cite{wang2024unigrasptransformer}, and visuo-tactile representations~\cite{yuan2024robot}. Here, \emph{unified} denotes a single generalist policy for grasp-and-lift manipulation across objects.

Our key contributions are summarized as follows:
\begin{itemize}
    \item A human-video-guided pipeline using bounded residual RL specialists to synthesize executable action--contact demonstrations, enabling the learning of a reference-free generalist policy.
    \item A four-bit contact representation combining spatial fingertip labels with a separate contact token for offline visuo-tactile policy learning, evaluated through observation and fusion comparisons.
    \item A complete sim-to-real implementation using 50 annotated human demonstrations and 10,000 simulated trajectories, evaluated on six seen and five unseen physical objects without real-robot demonstrations or policy fine-tuning.
\end{itemize}

\section{Related Work}
\label{sec:related}

\subsection{Learning Dexterous Manipulation from Human Videos}
Human videos support dexterous learning through demonstration translation~\cite{qin2022dexmv} or learned motion and interaction priors~\cite{shaw2023videodex,singh2024hop}. Annotated datasets such as DexYCB~\cite{chao2021dexycb} provide explicit hand--object geometry, but transferring these references to robots must account for embodiment and contact dynamics. This motivates reference-guided adaptation~\cite{chen2024object,zhao2024dexh2r,li2025maniptrans,liu2025dextrack}.

Simulation also connects geometric references to deployable policies. ViViDex~\cite{chen2024vividex} uses successful simulated rollouts to train a unified visual policy, while Video2Sim2Real~\cite{han2026video2sim2real} explores a video-to-simulation-to-reality pipeline. We follow this direction to generate offline demonstrations pairing robot action targets with robot-side contact observations.

\subsection{Specialist Distillation and Sensory Policy Learning}
Specialist-to-generalist learning separates object-specific control from deployment of a shared policy~\cite{wang2024unigrasptransformer}. Simulated demonstration augmentation~\cite{wang2024cyberdemo} and point-cloud representations for imitation learning~\cite{ze20243d} and dexterous control~\cite{qin2023dexpoint} provide complementary tools for this transition. We train a single ACT policy~\cite{zhao2023learning} on pooled specialist trajectories containing geometry and contact observations, keeping reference-guided adaptation in the data-generation stage.

\subsection{Visuo-Tactile Policy Learning}
Visuo-tactile learning combines geometric observations with local contact cues through spatial fusion~\cite{huang2024vitac}, cross-modal features~\cite{heng2025vitacformer}, or human tactile demonstrations~\cite{yu2023mimictouch}. Robot Synesthesia~\cite{yuan2024robot} combines labeled point clouds with binary tactile inputs for in-hand manipulation. We adopt a related representation principle for arm--hand grasp-and-lift manipulation, encoding fingertip contact through spatial point labels and a separate contact token.

Unlike approaches that rely on tactile measurements collected during human or robot demonstrations, UniDex-ViTac obtains robot-side contact observations during human-reference-guided simulated interaction. These observations are retained in the offline demonstration dataset and used to train a unified visuo-tactile policy that operates without human references or privileged object states at deployment.

\section{Method}
\label{sec:method}

\subsection{Problem Formulation}
\label{subsec:overview}
Given annotated human grasp-and-lift videos for $K$ training objects, we seek one policy $\pi_\theta(\mathbf{a}_t\mid\mathbf{P}_t,\mathbf{p}_t,\mathbf{c}_t)$ with deployable inputs. Here, $\mathbf{P}_t$ denotes the scene point cloud, $\mathbf{p}_t$ the robot proprioceptive state, and $\mathbf{c}_t$ the binary fingertip contact state. An object-specific specialist $\pi_k^{\mathrm{spec}}$ uses privileged simulated state and a human reference to generate demonstrations. Its rollouts provide paired sensory observations and final task-space targets for offline generalist learning.

\subsection{Human--Object Interaction References}
\label{subsec:references}
We use DexYCB~\cite{chao2021dexycb}, whose annotations include MANO~\cite{romero2022embodied} hand parameters and object poses. At time $t$, the human wrist pose $\mathbf w_t^H$, wrist-relative human fingertip positions $\mathbf h_t^H$, and object pose $\mathbf o_t^{\mathrm{ref}}$ form the human--object interaction (HOI) reference $\hat{\boldsymbol\tau}_t$. We map the human hand reference to a robot wrist target and four fingertip targets in the local wrist frame. The object trajectory supplies conditioning and a tracking objective for the specialist.

From the 20 objects in DexYCB, we retain five demonstrations for each of ten objects, giving 50 human demonstrations. Sequences with initial configurations unsuitable for the simulated tabletop setup, including unsupported or unstable placements, are excluded. The selected objects are master chef can, sugar box, tomato soup can, mustard bottle, pudding box, potted meat can, bleach cleanser, mug, power drill, and wood block.

\subsection{Object-Specific Residual RL}
\label{subsec:specialists}
Residual specialists adapt mapped references to the robot's kinematics and contact dynamics through bounded task-space corrections. We train one specialist per object with PPO~\cite{schulman2017proximal} in Isaac Lab~\cite{mittal2025isaac}, using 8,192 parallel environments per policy.

\paragraph{Privileged observation}
Both actor and critic receive $\mathbf{o}^{\mathrm{spec}}_t=(\mathbf{s}^{\mathrm{priv}}_t,\mathbf{p}_t,\mathbf{b}_t,\mathbf{F}_t,\hat{\boldsymbol{\tau}}_t,\mathbf{e}_t)$. Here $\mathbf{s}^{\mathrm{priv}}_t$ contains joint torques and velocities, object linear and angular velocities, and fingertip velocities. The proprioceptive state $\mathbf{p}_t$ is defined in~\eqref{eq:proprio_state}; $\mathbf{b}_t\in\{0,1\}^4$ contains binary fingertip contacts; and $\mathbf{F}_t$ contains continuous contact-force features. Object information $\mathbf{e}_t$ includes mass, friction, scale, and six-degree-of-freedom pose. Actor and critic are MLPs with hidden dimensions $[1024,1024,512,256]$ and ELU activations.

\paragraph{Residual accumulation and target construction}
The actor predicts a raw task-space residual command $\mathbf{u}_t$ at 20\,Hz. An exponential moving average smooths the command as $\tilde{\mathbf{u}}_t=\alpha\mathbf{u}_t+(1-\alpha)\tilde{\mathbf{u}}_{t-1}$, where $\alpha$ is the smoothing coefficient. The accumulated task-space residual $\Delta\mathbf{a}_t$ is then updated by
\begin{equation}
\begin{split}
    \Delta\mathbf{a}_t=\operatorname{clip}\big(&\Delta\mathbf{a}_{t-1}+s\tilde{\mathbf{u}}_t\Delta t,\;\\
    &\Delta\mathbf{a}_{\min},\Delta\mathbf{a}_{\max}\big),
\end{split}
    \label{eq:residual}
\end{equation}
where $s$ is the action scale and $\Delta t=0.05$\,s. The lower and upper bounds $\Delta\mathbf{a}_{\min}$ and $\Delta\mathbf{a}_{\max}$ limit cumulative corrections to $\pm0.1$\,m for wrist translation, $\pm40^\circ$ for wrist orientation, and $\pm0.05$\,m for wrist-frame fingertip positions.

The corrected target is obtained by adding $\Delta\mathbf{a}_t$ to the mapped video reference in task-space coordinates. Wrist rotation is accumulated in Euler-angle coordinates. The resulting wrist pose and fingertip targets are encoded as the generalist action in~\eqref{eq:action}.

\paragraph{Reward and training}
The reward combines object, fingertip, and wrist tracking with contact activation and motion penalties:
\begin{equation}
\begin{split}
    r_t={}&\omega_{\mathrm{obj}}r_{\mathrm{obj}}
       +\omega_{\mathrm{hand}}r_{\mathrm{hand}}
       +\omega_{\mathrm{wrist}}r_{\mathrm{wrist}}\\
       &+\omega_{\mathrm{contact}}r_{\mathrm{contact}}
       +\omega_{\mathrm{penalty}}r_{\mathrm{penalty}}.
\end{split}
    \label{eq:reward}
\end{equation}
The tracking terms use exponentials of negative positional and rotational errors relative to the references. The contact term $r_{\mathrm{contact}}=\norm{\mathbf{b}_t}_1$ counts active fingertip contacts. The penalty $r_{\mathrm{penalty}}=p_{\mathrm{col}}+p_{\mathrm{reg}}$ combines excessive fingertip and robot-body impact penalties with joint-velocity regularization, $p_{\mathrm{reg}}=-a_v\norm{\dot{\mathbf{q}}_t}_2$, where $\mathbf{q}_t$ collects the robot's joint positions. The $\omega$ coefficients weight the five reward terms.

Following the general principle of domain randomization~\cite{tobin2017domain}, we broaden the simulated training distribution. In our implementation, references are translated and rotated about the vertical axis through the object's initial position, and object physical properties and scale are randomized. Each specialist is trained for 1,000 policy updates. Training all ten specialists and collecting their rollouts takes approximately two days on one NVIDIA RTX 5090 GPU, excluding generalist training.

\subsection{Robot Demonstrations from Specialist Rollouts}
\label{subsec:dataset}
A trial succeeds when the robot grasps the object, lifts it by at least 20\,cm, and holds it for at least 3\,s. Before a candidate trajectory is admitted to the offline dataset, the corresponding specialist is re-evaluated in five verification rollouts with varied initial perturbations, and the candidate is retained only when all five rollouts succeed. The final dataset contains 10,000 retained trajectories, with 1,000 per object.

We pool the retained trajectories into the offline dataset $\mathcal{D}$. Each trajectory is a sequence of observation--action pairs,
\begin{equation}
    \tau^{(i)}=\{(\mathbf{s}^{(i)}_t,\mathbf{a}^{(i)}_t)\}_{t=0}^{T^{(i)}-1},
    \label{eq:trajectory}
\end{equation}
where $T^{(i)}$ is its length, $\mathbf{s}_t$ is the sensory observation, and $\mathbf{a}_t$ is the final corrected task-space target produced by the specialist. Robot-side contact is recorded in $\mathbf{s}_t$, and the target is stored in the action representation of~\eqref{eq:action}. Learning these final targets enables reference-free execution.

\subsection{Unified Visuo-Tactile Generalist}
\label{subsec:generalist}
\begin{figure}[t]
    \centering
    \includegraphics[width=\columnwidth]{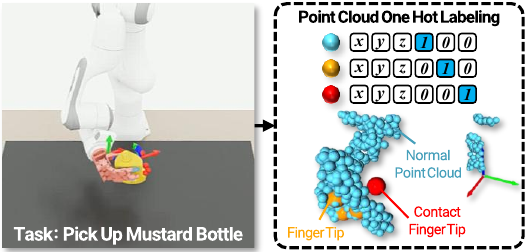}
    \caption{Point-cloud observation: 508 camera points and four FK fingertip positions, augmented with point-type and binary-contact labels. Fingertip positions provide spatial anchors for contact activation.}
    \label{fig:figure_2}
    \vspace{-0.4cm}
\end{figure}

\paragraph{Point cloud and contact labels}
The full visuo-tactile observation comprises a point cloud $\mathbf{P}_t$, proprioception $\mathbf{p}_t$, and binary fingertip contact $\mathbf{c}_t$:
\begin{equation}
    \mathbf{s}_t=\big(\mathbf{P}_t,\,\mathbf{p}_t,\,\mathbf{c}_t\big).
    \label{eq:observation}
\end{equation}
We sample 508 camera points by farthest point sampling (FPS) and append four FK fingertip positions. Each point stores its 3D position and a three-dimensional one-hot feature, giving $\mathbf{P}_t\in\mathbb{R}^{512\times6}$. The feature is $(1,0,0)$ for camera points, $(0,1,0)$ for non-contacting fingertips, and $(0,0,1)$ for contacting fingertips. All points are expressed in a common scene frame. The four-bit vector $\mathbf{c}_t\in\{0,1\}^4$ is also supplied as a separate input. Fingertip positions provide spatial anchors for contact activation (Fig.~\ref{fig:figure_2}).

\paragraph{Proprioception and actions}
The proprioceptive state concatenates the robot's kinematic features:
\begin{equation}
    \mathbf{p}_t=\operatorname{concat}\big(
    \mathbf{q}_{\mathrm{hand},t},\,
    \mathbf{q}_{\mathrm{arm},t},\,
    \mathbf{w}_t,\,
    \mathbf{f}_t\big)\in\mathbb{R}^{42},
    \label{eq:proprio_state}
\end{equation}
where $\mathbf{q}_{\mathrm{hand},t}\in\mathbb{R}^{16}$ and $\mathbf{q}_{\mathrm{arm},t}\in\mathbb{R}^{7}$ are the hand and arm joint positions. The wrist pose $\mathbf{w}_t\in\mathbb{R}^{7}$ contains position and quaternion, and $\mathbf{f}_t\in\mathbb{R}^{12}$ stacks the current positions of the four fingertips in the world frame.

The action specifies the desired wrist pose and wrist-relative fingertip targets:
\begin{equation}
    \mathbf{a}_t=\big(
    \mathbf{w}^{\mathrm{des}}_t,\,
    \mathbf{f}^{\mathrm{rel}}_t\big).
    \label{eq:action}
\end{equation}
Here $\mathbf{w}^{\mathrm{des}}_t\in\mathbb{R}^{9}$ contains the target wrist position and its continuous 6D rotation encoding~\cite{zhou2019continuity}. The vector $\mathbf{f}^{\mathrm{rel}}_t\in\mathbb{R}^{12}$ stacks the \emph{desired} positions of the four fingertips in the local wrist frame; $\mathrm{rel}$ denotes this coordinate frame. Thus, $\mathbf{a}_t$ has 21 components. A damped least-squares IK solver for the arm tracks the wrist target, and an analytical IK solver for the hand tracks the fingertip targets. Specialist demonstrations and generalist predictions share this action representation.

\begin{figure}[t]
    \centering
    \includegraphics[width=\columnwidth]{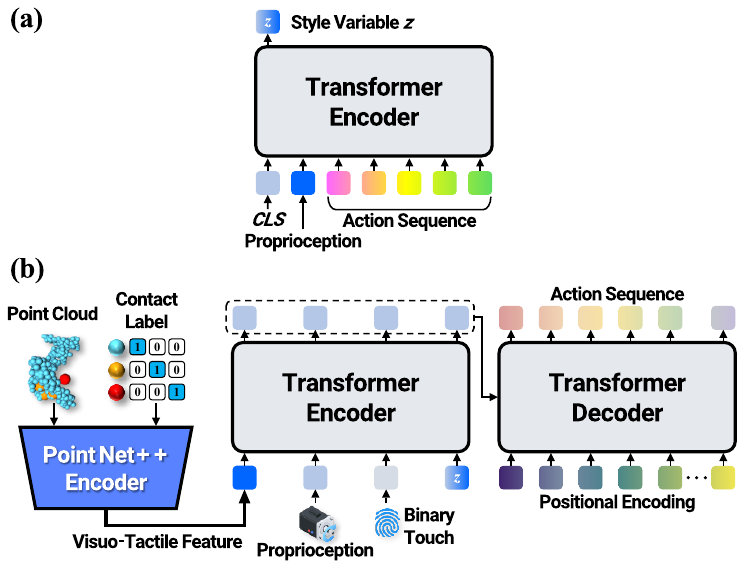}
    \caption{ACT-based generalist. (a) Training-time CVAE encoder. (b) Point-cloud, proprioceptive, and contact tokens condition a Transformer that predicts a 30-step action chunk.}
    \label{fig:figure_5}
    \vspace{-0.4cm}
\end{figure}

\paragraph{Action-chunking architecture}
We adapt ACT~\cite{zhao2023learning} by replacing its image pathway with PointNet++~\cite{qi2017pointnet++} and tactile encoders (Fig.~\ref{fig:figure_5}). PointNet++ encodes $\mathbf P_t$, a linear layer projects $\mathbf p_t$, and an MLP encodes $\mathbf c_t$. The three observation tokens and a conditional variational autoencoder (CVAE) latent $z$ condition the action predictor.

The Transformer predicts a 30-step action chunk from the encoded observations and latent. During training, the CVAE encoder infers the latent distribution from proprioception and the demonstrated action chunk. Action reconstruction and KL regularization toward a standard normal latent prior jointly train the encoder and policy. Both contact paths encode the same four bits as conditioning inputs. At deployment, the CVAE encoder is discarded, the latent is fixed to the prior mean, and temporal ensembling~\cite{zhao2023learning} combines time-aligned predictions from overlapping chunks into the current target command.

\subsection{A Four-Bit Tactile Interface}
\label{subsec:tactile}
One bit per fingertip provides a shared contact interface across simulation and hardware. In simulation, $b_{t,j}$ is set to 1 when the scalar total-contact-force measure $F^{\mathrm{pad}}_{t,j}$ on fingertip pad $j$ exceeds 1.0\,N, and to 0 otherwise. Each bit records above-threshold contact with the object or environment. These four bits are stored as $\mathbf{c}_t=\mathbf{b}_t$ in generalist demonstrations.

\begin{figure}[t]
    \centering
    \includegraphics[width=0.9\columnwidth]{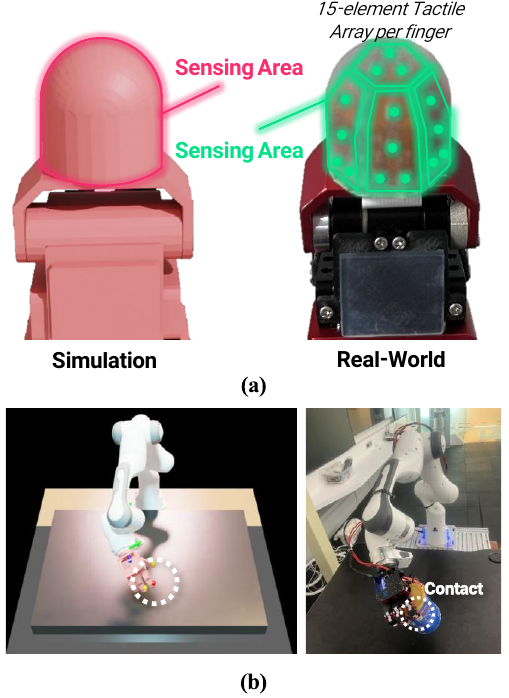}
    \caption{Binary tactile sensing. (a) Simulated fingertip-pad and real sensing regions; the real sensitive area can extend beyond the sensor footprint. (b) Real contact activations visualized in the digital twin.}
    \label{fig:figure_6}
    \vspace{-0.4cm}
\end{figure}

On hardware, each finger carries barometric pressure sensors. Each sensor is calibrated by applying a constant 1.0\,N force for 10\,s and using the mean raw output as its threshold. The contact bit for a fingertip is set to 1 whenever at least one of its sensors exceeds its own threshold, and to 0 otherwise. The reported raw output range is 0--8192. The real sensing region can extend beyond the sensor footprint (Fig.~\ref{fig:figure_6}).

\section{Experiments and Results}
\label{sec:experiments}
We examine \textbf{RQ1}, specialist data generation and the transition to a generalist; \textbf{RQ2}, generalist observation configurations; \textbf{RQ3}, contact-fusion design; and \textbf{RQ4}, physical seen- and unseen-object deployment.

\subsection{Experimental Setup}
\label{subsec:setup}
\paragraph{Simulation and evaluation}
Training and evaluation use Isaac Lab and the ten selected objects, with the arm--hand model used for deployment. Initial positions are perturbed by up to $\pm15$\,cm in the $xy$-plane and yaw by up to $\pm15^\circ$. A simulated depth camera is placed at the physical camera pose. Success uses the 20\,cm lift and 3\,s hold criterion of Sec.~\ref{subsec:dataset}.

Both specialist and generalist simulation evaluations use five \emph{evaluation seeds}, with 100 trials per seed and object, for 500 evaluation trials per object. We report pooled per-object success rates and their unweighted mean across objects (Table~\ref{tab:table_4.1}).

\paragraph{Compared configurations}
All generalists are trained on the same 10,000 specialist trajectories. State-based ACT, Diffusion Policy (DP)~\cite{chi2023diffusion}, and BC-Transformer (BC-T)~\cite{mandlekar2021matters} use proprioception, one-hot object identity, and ground-truth object pose. State+Contact adds binary contact to state-based ACT. For the point-cloud-only configuration (PCD Only), the point-cloud input contains only camera-observed geometry: no contact token, contact-dependent one-hot features, or appended contact-labeled fingertip points are supplied. PCD+Contact uses the full representation of Sec.~\ref{subsec:generalist}, including the four labeled FK fingertip points and separate token. In the fusion study, One-hot Only enables the spatial fingertip-label path while omitting the separate contact token, Token Only uses the contact-token path without contact-dependent point labels, and PCD+Contact uses both paths.

The state-based baselines use 5,000 training epochs, with validation-loss plateauing reported in the training runs; the visuo-tactile model uses 1,500 epochs. Cross-architecture results therefore compare complete model, observation, and training configurations; the ACT variants assess observation design.

\paragraph{Physical platform and policy execution}
The robot uses a Franka Emika Panda arm, a 16-DoF hand, and the tactile interface of Sec.~\ref{subsec:tactile}. An Intel RealSense L515 supplies point clouds, transformed to the robot base frame and cropped to the workspace. The same observation definitions are used in simulation and hardware. Deployment includes camera alignment and per-sensor tactile calibration.

The nominal target-command rate is 20\,Hz, with a 30-action prediction horizon (1.5\,s). The arm/hand IK interface is shared across simulation and hardware. We evaluate six seen and five unseen objects with ten trials per object \emph{per policy}: 110 per policy and 220 in total. Placement and yaw are randomized.

\subsection{From Specialists to One Generalist (RQ1)}
\label{subsec:rl_results}
\begin{figure}[t]
    \centering
    \includegraphics[width=0.8\columnwidth]{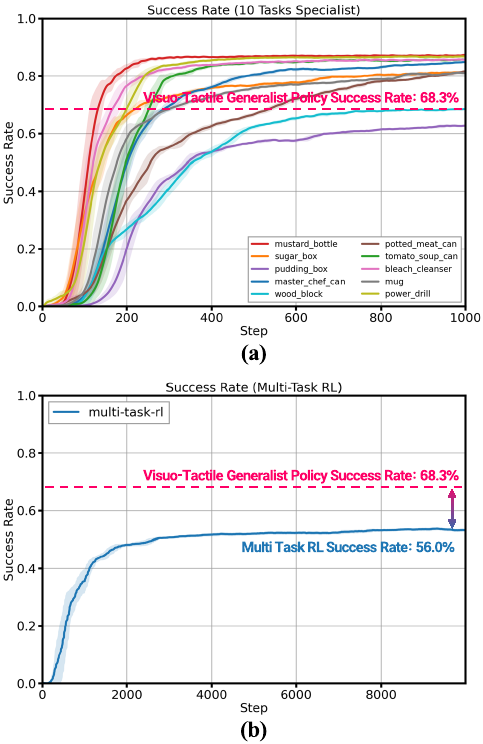}
    \caption{RL learning curves for (a) object-specific specialists, trained for 1{,}000 updates each, and (b) Multi-Task RL, trained for 10{,}000 updates. The Multi-Task RL run ends at 56.0\%. The dashed 68.3\% line marks the final generalist evaluation result.}
    \label{fig:figure_7}
    \vspace{-0.4cm}
\end{figure}

Fig.~\ref{fig:figure_7} shows the ten specialist learning curves and the Multi-Task RL baseline. Each specialist uses 8,192 environments for 1,000 updates; Multi-Task RL uses 8,192 environments across all ten objects for 10,000 updates with the same specified observation, residual-action, reward, and network settings. Successful specialist rollouts form the balanced offline dataset for generalist learning.

The generalist achieves 68.3\% in final evaluation, above the 56.0\% training-time endpoint of Multi-Task RL. Table~\ref{tab:table_4.1} compares final specialist and generalist performance using 500 evaluation trials per object, with macro-averages of 94.3\% and 68.3\%, respectively. The 26.0-point gap compares object-specific specialists using privileged state and human references with a single generalist using deployable sensory inputs.

\begin{table}[!t]
    \renewcommand{\arraystretch}{1.1}
    \centering
    \caption{Per-object success rates (\%) for object-specific Specialist RL and the unified Generalist Policy. Both columns aggregate 500 evaluation trials per object (five evaluation seeds, 100 trials per seed). Gap is specialist minus generalist in percentage points; the final row is the unweighted mean over objects.}
    \label{tab:table_4.1}
    \footnotesize
    \setlength{\tabcolsep}{3pt}
    \begin{tabular*}{\linewidth}{@{\extracolsep{\fill}}lrrr@{}}
        \toprule
        \textbf{Object} &
        \makecell[r]{\textbf{Specialist}\\\textbf{RL}} &
        \makecell[r]{\textbf{Generalist}\\\textbf{Policy}} &
        \makecell[r]{\textbf{Gap}\\\textbf{(pp)}} \\
        \midrule
        \texttt{002\_master\_chef\_can} & 98.2 & 68.0 & 30.2 \\
        \texttt{004\_sugar\_box}        & 89.8 & 56.0 & 33.8 \\
        \texttt{005\_tomato\_soup\_can} & 98.0 & 65.8 & 32.2 \\
        \texttt{006\_mustard\_bottle}   & 99.8 & 88.0 & 11.8 \\
        \texttt{008\_pudding\_box}      & 76.0 & 48.0 & 28.0 \\
        \texttt{010\_potted\_meat\_can} & 95.8 & 55.0 & 40.8 \\
        \texttt{021\_bleach\_cleanser}  & 97.6 & 94.0 & 3.6  \\
        \texttt{025\_mug}              & 94.8 & 48.0 & 46.8 \\
        \texttt{035\_power\_drill}      & 99.4 & 94.0 & 5.4  \\
        \texttt{036\_wood\_block}       & 93.2 & 66.0 & 27.2 \\
        \midrule[\heavyrulewidth]
        \textbf{Macro-average} & \textbf{94.3} & \textbf{68.3} & \textbf{26.0} \\
        \bottomrule
    \end{tabular*}
\end{table}

\begin{figure}[t]
    \centering
    \includegraphics[width=\columnwidth]{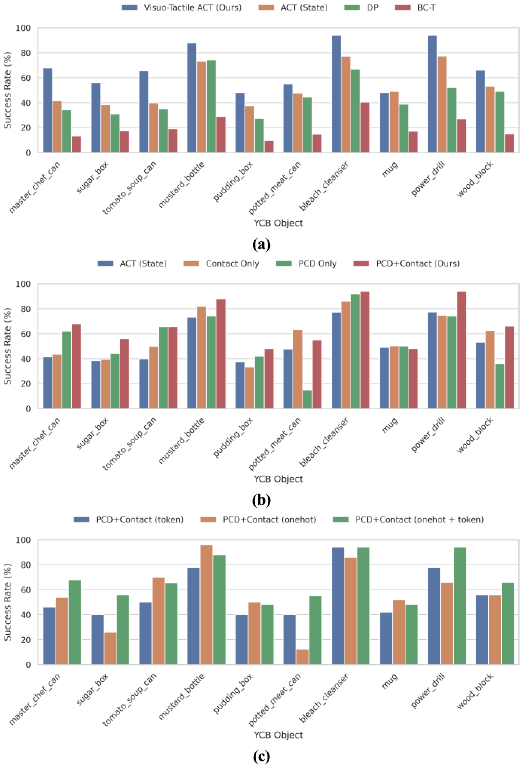}
    \caption{Simulation results on ten YCB objects. Plot labels correspond to the text as follows: \emph{Visuo-Tactile ACT (Ours)} = PCD+Contact, \emph{Contact Only} = State+Contact, and \emph{token/onehot/onehot+token} = Token Only/One-hot Only/PCD+Contact.}
    \label{fig:figure_3}
    \vspace{-0.4cm}
\end{figure}

\subsection{Observation Configurations (RQ2)}
\label{subsec:observation_results}
PCD+Contact achieves 68.3\% macro-average success, compared with 55.5\% for PCD Only, a difference of 12.8 percentage points (Fig.~\ref{fig:figure_3}(a,b)). State+Contact achieves 58.4\% versus 53.4\% for State, a 5.0-point difference.

The 12.8-point difference compares camera-derived point clouds with the full representation augmented by labeled FK fingertip points and a separate contact token. The fusion study below examines the spatial-label and token paths within this representation.

The state-based DP and BC-T baselines achieve 45.2\% and 20.2\%, respectively. Their privileged object-state inputs and policy architectures differ from the proposed sensory policy, providing configuration-level comparisons under the training schedules of Sec.~\ref{subsec:setup}.

\subsection{Contact-Fusion Design (RQ3)}
\label{subsec:fusion_results}
Fig.~\ref{fig:figure_3}(c) compares One-hot Only, Token Only, and the dual-path PCD+Contact configuration. The plotted per-object results correspond to macro-averages of 56.8\% for One-hot Only, 56.4\% for Token Only, and 68.3\% for PCD+Contact, compared with 55.5\% for PCD Only. Thus, the single-path variants improve the macro-average by only 1.3 and 0.9 percentage points, whereas using both paths yields the full 12.8-point increase. A possible explanation for this non-additive pattern is complementary conditioning from spatial fingertip labels and the global contact token. The ranking also varies by object: One-hot Only is higher on mustard bottle, whereas PCD+Contact is higher on potted meat can and power drill.

\begin{figure}[t]
    \centering
    \includegraphics[width=0.9\columnwidth]{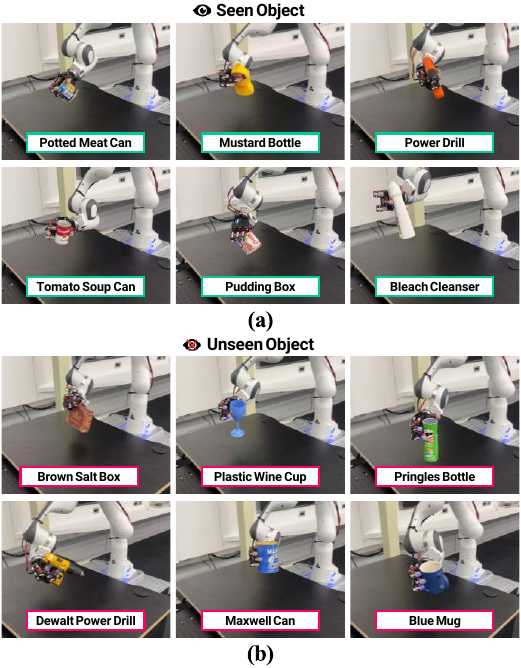}
    \caption{Examples of real-world rollouts on (a) seen and (b) unseen objects. The objects included in the quantitative evaluation are listed in Table~\ref{tab:table_1}.}
    \label{fig:figure_4}
    \vspace{-0.4cm}
\end{figure}

\begin{table}[!t]
\centering
\caption{Real-world success over ten trials per object per policy for PCD Only and PCD+Contact.}
\label{tab:table_1}
\small
\setlength{\tabcolsep}{3pt}
\begin{tabular}{llcc}
\toprule
\textbf{Category} & \textbf{Object Name} & \makecell{\textbf{PCD}\\\textbf{Only}} & \makecell{\textbf{PCD+}\\\textbf{Contact}} \\
\midrule
\multirow{6}{*}{\textbf{Seen}}
 & \texttt{bleach\_cleanser} & 10/10 & 9/10 \\
 & \texttt{mustard\_bottle} & 7/10 & 7/10 \\
 & \texttt{power\_drill} & 7/10 & 8/10 \\
 & \texttt{potted\_meat\_can} & 4/10 & 7/10 \\
 & \texttt{tomato\_soup\_can} & 3/10 & 5/10 \\
 & \texttt{pudding\_box} & 2/10 & 6/10 \\
\cmidrule(lr){2-4}
 & \textbf{Average} & \textbf{55.0\%} & \textbf{70.0\%} \\
\midrule
\multirow{5}{*}{\textbf{Unseen}}
 & \texttt{maxwell\_coffee\_can} & 9/10 & 9/10 \\
 & \texttt{brown\_salt\_box} & 2/10 & 4/10 \\
 & \texttt{plastic\_wine\_cup} & 7/10 & 6/10 \\
 & \texttt{blue\_mug} & 0/10 & 2/10 \\
 & \texttt{pringles\_bottle} & 9/10 & 10/10 \\
\cmidrule(lr){2-4}
 & \textbf{Average} & \textbf{54.0\%} & \textbf{62.0\%} \\
\midrule
\multicolumn{2}{l}{\textbf{Overall Average}} & \textbf{54.5\%} & \textbf{66.4\%} \\
\bottomrule
\end{tabular}
\vspace{-0.3cm}
\end{table}

\subsection{Transfer to Physical Objects (RQ4)}
\label{subsec:real_results}
Table~\ref{tab:table_1} reports all quantitative physical trials for PCD Only and PCD+Contact. PCD+Contact succeeds in 42/60 seen-object and 31/50 unseen-object trials, versus 33/60 and 27/50 for PCD Only. Overall, the counts are 73/110 (66.4\%) and 60/110 (54.5\%), a difference of 13 successes or 11.8 percentage points before rounding.

The largest seen-object differences favor contact on pudding box (four successes), potted meat can (three), and tomato soup can (two). Brown salt box and blue mug each gain two successes. PCD+Contact records one fewer success on bleach cleanser and plastic wine cup, highlighting object-dependent outcomes.

Both configurations transfer to unseen objects within the same grasp-and-lift skill. PCD Only achieves 27/50 successes, while PCD+Contact reaches 31/50 using the same trained generalist without object-specific retraining at deployment.

Fig.~\ref{fig:figure_4} uses shortened display labels: \emph{Maxwell Can} corresponds to \texttt{maxwell\_coffee\_can} in Table~\ref{tab:table_1}, while the DeWalt drill is a qualitative example outside the quantitative set.

\section{Conclusion and Limitations}
\label{sec:conclusion}

UniDex-ViTac connects annotated human videos to a deployable visuo-tactile grasping policy through simulated residual adaptation and offline learning. Object-specific specialists use human references to generate robot demonstrations paired with fingertip contact observations, from which a single generalist learns to operate without human references or privileged object identity and pose. The contact-augmented configuration achieves 68.3\% macro-average success in simulation, compared with 55.5\% for the point-cloud-only baseline, and 73/110 physical successes compared with 60/110 without real-robot demonstrations or policy fine-tuning. These correspond to observed improvements of 12.8 and 11.8 percentage points, respectively. The results support the feasibility of retaining robot-side contact observations generated in simulation when learning a unified sensory policy from human-video-guided interactions.

Our current tactile interface is intentionally sparse, using only four binary fingertip contact signals and therefore discarding force magnitude and detailed contact-location information. In addition, the simulated fingertip-pad contact region and the effective sensing region of the physical tactile sensors are not spatially identical, introducing a sim-to-real discrepancy in contact activation. Finally, the current evaluation focuses on a single grasp-and-lift skill. Future work will investigate richer tactile representations, improved sim-to-real alignment of contact sensing, and extension of the framework to longer-horizon, multi-stage, and more complex contact-rich manipulation tasks.

\bibliographystyle{Styles/jabbrv_ieeetr}


\bibliography{References/references}

\end{document}